\documentclass[letterpaper, 10 pt,conference]{ieeeconf}
\IEEEoverridecommandlockouts 
\usepackage{cite}

\usepackage{amsmath,amssymb,amsfonts}
\usepackage{graphicx}
\usepackage{textcomp}
\usepackage[table,xcdraw]{xcolor}
\usepackage{subcaption}
\usepackage{authblk}
\let\labelindent\relax
\usepackage{enumitem}
\usepackage{hyperref}
\usepackage{breqn}
\usepackage[linesnumbered,ruled,lined]{algorithm2e}
\usepackage{mdframed}
\usepackage{array}
\usepackage{makecell}
\usepackage{tabularray}
\usepackage{tcolorbox}
\usepackage{svg}
\usepackage{multirow}
\usepackage{nicematrix}
\usepackage{wrapfig}
\usepackage{booktabs}
\usepackage{diagbox}
\usepackage{adjustbox}

\allowdisplaybreaks

\SetKwComment{Comment}{/* }{ */}

\begin{document}

\title{$\pi$-MPPI: A Projection-based Model Predictive Path Integral Scheme for Smooth Optimal Control of Fixed-Wing Aerial Vehicles   
}

\author{
Edvin Martin Andrejev$^{1,2}$,
Amith Manoharan$^{1,2}$,
Karl-Eerik Unt$^{2}$,
Arun Kumar Singh$^{1,2}$

\thanks{$^1$ are with the Institute of Technology, University of
Tartu, Tartu, Estonia. }
\thanks{$^2$ are with the Estonian Aviation Academy, Tartu, Estonia.}
\thanks{This research was in part supported by grant PSG753 from the Estonian Research Council and SekMO program (2021-2027.1.01.23-0419, 2021-2027.1.01.24-0624, 2021-2027.1.01.23-0385) co-funded by European Union.
Emails: aks1812@gmail.com. Code: https://github.com/edvinmandrejev/Pi-MPPI.git}
}

\maketitle

\begin{abstract}

Model Predictive Path Integral (MPPI) is a popular sampling-based Model Predictive Control (MPC) algorithm for nonlinear systems. It optimizes trajectories by sampling control sequences and averaging them. However, a key issue with MPPI is the non-smoothness of the optimal control sequence, leading to oscillations in systems like fixed-wing aerial vehicles (FWVs). Existing solutions use post-hoc smoothing, which fails to bound control derivatives. This paper introduces a new approach: we add a projection filter $\pi$ to minimally correct control samples, ensuring bounds on control magnitude and higher-order derivatives. The filtered samples are then averaged using MPPI, leading to our $\pi$-MPPI approach. We minimize computational overhead by using a neural accelerated custom optimizer for the projection filter. $\pi$-MPPI offers a simple way to achieve arbitrary smoothness in control sequences. While we focus on FWVs, this projection filter can be integrated into any MPPI pipeline. Applied to FWVs, $\pi$-MPPI is easier to tune than the baseline, resulting in smoother, more robust performance.

\end{abstract}


\section{Introduction}
Model Predictive Path Integral (MPPI) \cite{williams2017model} is a popular gradient-free, sampling-based approach for computing optimal trajectories. The core idea of MPPI is to sample a sequence of control inputs from a Gaussian distribution and then approximate the optimal control as a weighted combination of the samples. The weights are determined by evaluating the cost of the forward rollouts (trajectories) of the system. Despite its versatility, it has one key limitation: the optimal controls resulting from MPPI are highly jerky. This, in turn, prevents its application to systems like fixed-wing aerial vehicles (FWV), where jerky control can lead to sharp oscillations in the actuation surfaces, which in turn can destabilize the vehicle.

Existing works address the non-smoothness issue by post-hoc smoothing \cite{mohamed2022autonomous} through the Savitzky–Golay filter. The smoothing can also be assisted by tuning the covariance of the control sampling distribution \cite{mohamed2022autonomous}. Recently, \cite{kim2022smooth} introduced the idea of ensuring smoothness by artificially increasing the degree of the systems. Specifically, \cite{kim2022smooth} performs MPPI in the space of control derivatives and treats controls as additional state variables, allowing them to include additional control penalties. However, 
such an approach can introduce additional challenges from the perspective of appropriately tuning the cost function.



\noindent \textit{Algorithmic Contributions:} We introduce $\pi$-MPPI, a variant of MPPI ensuring smooth optimal controls. By embedding a projection filter $\pi$ that minimally adjusts samples from a Gaussian distribution, we enforce higher-order derivative constraints on control inputs. The MPPI is then computed by averaging over these filtered sequences with an updated rule.

The projection filter $\pi$ is solved via quadratic programming (QP), and we present a fast, parallelizable custom solver for it. This maintains real-time applicability, which is critical for agile systems like FWVs. Additionally, we propose self-supervised learning to train a neural network for warm-starting the QP solver, making $\pi$-MPPI as efficient as the baseline variant. 

\noindent \textit{State-of-the-Art Performance:} $\pi$-MPPI avoids control bound penalties in the cost function, simplifying tuning, and can handle a higher control covariance without violating actuation bounds. We validate its performance on two FWV benchmarks: one requiring 3D target encirclement with obstacle avoidance and another involving terrain-following while avoiding crashes \cite{kyriakis2018terrain}. Compared to baseline variants, $\pi$-MPPI shows improved success rates, better target proximity, and fewer violations of control and its derivatives.
\vspace{-0.5cm}
\section{Problem Formulation and Baseline MPPI}
\vspace{-0.1cm}
\subsubsection*{Symbols and Notation} The transpose of a vector or matrix is defined using superscript $\intercal$. Matrices are written using uppercase bold font letters. Lowercase italic letters represent scalars, and the bold font upright variants represent vectors. At some places in this paper, we will use a construction wherein a vector $\mathbf{x}$ is formed by stacking $\mathbf{x}_k$ at different time-step $k$.  
\vspace{-0.3cm}
\subsection{Model of the aircraft}
\noindent We use a reduced-order version of the kinematic 3D Dubins model of fixed-wing aircraft, assuming that the values of the angle of attack and the sideslip are small \cite{beard2012small}. The aircraft states are defined as positions along north, east, and down directions, $p_n, p_e, p_d$, and yaw angle, $\psi$. 
\begin{align}
    \dot{p}_{n} &= v \cos{\psi}\cos{\theta}, \, \label{eqn:n}
    \dot{p}_{e} = v \sin{\psi}\cos{\theta},  \,
    \dot{p}_{d} = -v \sin{\theta}, \\
    \dot{\psi} &= \frac{\sin{\phi}}{\cos{\theta}} q + \frac{\cos{\phi}}{\cos{\theta}} r, \,
    q = \frac{\dot{\theta}+r\sin{\phi}}{\cos{\phi}},\,
    r = \frac{g}{v}\sin{\phi}\cos{\theta}, \nonumber
\end{align}    
where $(v, \phi, \theta)$ are the forward speed, roll, and pitch angles of the FWV, respectively. The variables $q$ and $r$ are the angular velocities about the pitch and yaw axis, and $g$ is the acceleration due to gravity.

\subsection{Problem Formulation}
\noindent We consider a generic optimization problem of the following form:
\begin{subequations}
\begin{align}
    \min \sum_{k=0}^{K-1} &\left( c(\mathbf{x}_k)+\frac{1}{2}\mathbf{u}_k^\intercal\mathbf{R}\mathbf{u}_k \right)+c_K(\mathbf{x}_K), \label{cost_fun}\\
    &\text{s.t. } \mathbf{x}_{k+1} = \mathbf{f}_{dyn}(\mathbf{x}_k,\mathbf{u}_k), \label{dyn_model}\\
    & \qquad ^{(j)}\mathbf{u}_0 = \ ^{(j)}\mathbf{u}_{\mathrm{init}}, \label{boundary_cond}\\
     & \qquad ^{(j)}\mathbf{u}_{\mathrm{min}} \leq {^{(j)}}\mathbf{u}_k \leq {^{(j)}}\mathbf{u}_{\mathrm{max}}, \label{control_bound}
\end{align}    
\end{subequations}

\noindent where the left superscript refers to the $j$\textsuperscript{th} order derivative of the control input $\mathbf{u}$. For FWV, we consider $\mathbf{u}_k = \begin{bmatrix}
 v_k & \phi_k & \theta_k   
\end{bmatrix}$ and note that $\dot{\theta}_k$ can be easily derived from $\theta_k$ at different time-steps. 

The state-dependent running and terminal costs $c$ and $c_K$, respectively, in \eqref{cost_fun}, can be arbitrary, non-linear, and non-convex. The dynamics model $\mathbf{f}_{dyn}$ in \eqref{dyn_model} is simply a compact representation of  \eqref{eqn:n} with $\mathbf{x}_k = \begin{bmatrix}
    p_{n , k} & p_{e, k} & p_{d, k} & \psi_k
\end{bmatrix}$. The constraint \eqref{boundary_cond} enforces the boundary conditions on the derivatives of the control input up to the  $j$\textsuperscript{th} order. Finally, \eqref{control_bound} bounds the magnitude of the control input and their derivatives. Throughout this paper, we consider $j = \{0, 1, 2\}$.

\newtheorem{remark}{Remark}\label{rem_1}
\begin{remark}
   An MPC pipeline consists of repeatedly solving \eqref{cost_fun}-\eqref{control_bound} from the current initial state $\mathbf{x}_0$.
\end{remark}





\subsection{Baseline MPPI Algorithm (MPPIwSGF)}

\noindent We consider the algorithm presented in \cite{williams2018information} as the baseline MPPI variant. We consider an MPC setting where $\mathbf{u}_{i, k}$ will be the $k^{th}$ step control input at the $i^{th}$ MPC iteration. MPPI provides a closed-form formula for the $(i+1)^{th}$ iteration in terms of $\mathbf{u}_{i, k}$ . Let $\boldsymbol{\epsilon}_k \sim \mathcal{N}(0,\boldsymbol{\Sigma_k})$ and $\boldsymbol{\epsilon}_k^m$ be the $m^{th}$ sample drawn from the distribution, using which, we can define a perturbed control input $\boldsymbol{\nu}_k^m$ as 
\begin{align}
 \boldsymbol{\nu}_k^m = \mathbf{u}_{i, k} + \boldsymbol{\epsilon}_k^m.  
 \label{pert_control}
\end{align}

\noindent The key idea in MPPI is to use $\boldsymbol{\nu}_k^m$ to forward simulate the FWV using the discrete-time dynamics \eqref{dyn_model}, resulting in a state trajectory distribution. Starting from an initial state, $\mathbf{x}_0$, let $\mathbf{x}_k^m$ represent the state at time-step $k$, obtained using $\boldsymbol{\nu}_k^m$. We can define the cost associated with the $m^{th}$ rollout in the following manner,
\begin{align}
    s^m = \sum_{k=0}^{K-1}c(\mathbf{x}_k^m) + c_K(\mathbf{x}_K^m). \label{eqn:aug_cost}
\end{align}

\noindent The next step is to assign weights $w^m$ to each trajectory as
\begin{equation}
    w^m = \frac{1}{\eta}\exp{\left( -\frac{1}{\sigma}\left( s^m + \gamma\sum_{k=0}^{K-1}\mathbf{u}_{i, k}^{\intercal}\boldsymbol{\Sigma}_k^{-1}\boldsymbol{\nu}_k^m-\rho \right) \right)}, \label{eqn:weight}
\end{equation}
where
\begin{equation}
    \eta = \sum_{m=1}^M\exp{\left( -\frac{1}{\sigma}\left( s^m + \gamma\sum_{k=0}^{K-1}\mathbf{u}_{i, k}^{\intercal}\boldsymbol{\Sigma}_k^{-1}\boldsymbol{\nu}_k^m-\rho \right) \right)},
\end{equation}
\begin{equation}
    \rho = \min_{m=1\ldots M} \left( s^m+\gamma\sum_{k=0}^{K-1}\mathbf{u}_{i, k}^{\intercal}\boldsymbol{\Sigma}_k^{-1}\boldsymbol{\nu}_k^m \right),
\end{equation}
$M$ is the total number of sampled trajectories (rollouts), $K$ is the total timesteps, and $\sigma,\gamma$ are tunable parameters, with $\gamma=\sigma(1-\alpha)$, and $0<\alpha<1$. When the value of $\sigma$ increases, the trajectories get weighted equally. When $\sigma$ decreases, the lower-cost trajectories get more weightage. The term $\rho$ is included for numerical stability during implementation. 
The final step in MPPI is to calculate the optimal controls for the next iteration using the weighted average of the sampled controls across all trajectories as 
\begin{align}
    \mathbf{u}_{i+1, k} = \mathbf{u}_{i, k} + \sum_{m=1}^{M} w^m \boldsymbol{\epsilon}_k^m. \label{eqn:mean_u}
\end{align}

\noindent As mentioned earlier, a key challenge in MPPI is that optimal control resulting from \eqref{eqn:mean_u} is typically very jerky. 
Hence, a post-hoc smoothing is often required. In particular, a Savitzky–Golay filter (SGF) is often used to smooth the control sequence. Hence, from here on, we refer to this method as MPPIwSGF. The baseline MPPIwSGF can enforce just the control bounds by clipping $\boldsymbol{\nu}_k^m$. However, the bounds need not be preserved by the SGF filtering process. In the next section, we propose an approach that uses a projection filter $\pi$ for bounding the control and its derivatives, thus resulting in a highly smooth control profile. We refer to our method as the $\pi$-MPPI (Alg. \ref{alg:one}).

\begin{algorithm}[h]
\SetKwInput{KwData}{Input}
\caption{$\pi$-MPPI\\\scriptsize{The main differences with the baseline MPPIwSGF are highlighted in \textcolor{teal}{blue}.}}\label{alg:one}
\small
\KwData{Dynamics $\mathbf{f}_{dyn}$, MPPI parameters $\sigma,\gamma,\alpha$, no. of timesteps $K$, no. of rollouts $M$, mean input $\mathbf{u}_{i, k}$ from the previous MPC iteration $i$, covariance $\boldsymbol{\Sigma}_k$, initial state $\mathbf{x}_0$, initial boundary values for the $j$\textsuperscript{th} derivatives of control $^{(j)}\mathbf{u}_{\mathrm{init}}$}
\For{$m = 1 \hdots M$}{
\For{$k = 0 \hdots K-1$}{
$\boldsymbol{\epsilon}_k^m \sim \mathcal{N}(0,\boldsymbol{\Sigma}_k)$\quad //Sample the noise\\
$\boldsymbol{\nu}_k^m = \mathbf{u}_{i,k} + \boldsymbol{\epsilon}_k^m$\quad //Random control samples\\
}
Define $\boldsymbol{\nu}^m=\left[ \boldsymbol{\nu}_0^m,\boldsymbol{\nu}_1^m,\ldots,\boldsymbol{\nu}_{K-1}^m \right]$\\
\textcolor{teal}{Compute the projection for all the samples
    $\overline{\boldsymbol{\nu}}^m = \pi(\boldsymbol{\nu}^m, \, ^{(j)}\mathbf{u}_{\mathrm{init}}$) }
}
\For{$k = 0 \hdots K-1$}{
    \textcolor{teal}{$\boldsymbol{\overline{u}}_{i,k}$ = Mean$\left(\left[ \boldsymbol{\overline{\nu}}_k^1,\boldsymbol{\overline{\nu}}_k^2,\ldots,\boldsymbol{\overline{\nu}}_{k}^M \right]\right)$\quad//Update mean}\\ 
    \textcolor{teal}{$\overline{\boldsymbol{\Sigma}}_k$ \quad = Cov$\left(\left[ \boldsymbol{\overline{\nu}}_k^1,\boldsymbol{\overline{\nu}}_k^2,\ldots,\boldsymbol{\overline{\nu}}_{k}^M \right]\right)$\quad//Update cov.}\\
    }
\For{$m = 1 \hdots M$}{
\For{$k = 0 \hdots K-1$}{
    \textcolor{teal}{$\overline{\boldsymbol{\epsilon}}_k^m$ = $\overline{\boldsymbol{\nu}}_k^m-\overline{\mathbf{u}}_{i,k}$\quad //Update noise}\\
    Compute trajectory rollouts $\mathbf{x}_k^m$ using equation \eqref{dyn_model}}
\textcolor{teal}{Compute the cost $s^m= \sum_k c(\mathbf{x}_k^m,\overline{\boldsymbol{\nu}}_k^m)+c_K(\mathbf{x}_K^m)$}\\
\textcolor{teal}{Compute the weights $w^m$ using equation \eqref{eqn:weight} with the cost $s^m$ computed from line 18, $\overline{\mathbf{u}}_{i,k},\overline{\boldsymbol{\Sigma}}_k$, and $\overline{\boldsymbol{\nu}}_k^m$}
}
\For{$k = 0 \hdots K-1$}{
\textcolor{teal}{Compute the weighted average control $\mathbf{u}_{i+1,k} = \overline{\mathbf{u}}_{i,k} + \sum_{m=1}^{M} w^m \overline{\boldsymbol{\epsilon}}_k^m$ }
}
Define $\mathbf{u}_{i+1}=\left[ \mathbf{u}_{i+1,0},\mathbf{u}_{i+1,1},\ldots,\mathbf{u}_{i+1,K} \right]$\\
\textcolor{teal}{Compute the projection for the optimal control \qquad \qquad $\mathbf{u}_{i+1}^*$ = $\pi(\mathbf{u}_{i+1},\,^{(j)}\mathbf{u}_{\mathrm{init}})$}\\
$\mathbf{u}_{i+1,k} \leftarrow \mathbf{u}_{i+1}^*$ \quad //\textcolor{teal}{Update the mean for next MPC iteration}   
\normalsize
\end{algorithm}

\section{Main Algorithmic Results}
This section first presents an overview of $\pi$-MPPI, followed by our approach for constructing the projection filter in a computationally efficient manner.

\subsection{Overview of Projection based MPPI ($\pi$-MPPI)}

\noindent Alg.~\ref{alg:one} summarizes $\pi$-MPPI, highlighting differences from MPPIwSGF in \textcolor{teal}{blue}. As in the baseline MPPIwSGF, lines 3-4 sample $\boldsymbol{\epsilon}_k^m$ from a zero-mean Gaussian to construct $\boldsymbol{\nu}_k^m$. However, instead of using $\boldsymbol{\nu}_k^m$ directly for rollouts, we first apply a projection filter (line 7) to obtain $\overline{\boldsymbol{\nu}}_k^m$.

Since this filter involves non-linear operations, $\overline{\boldsymbol{\nu}}_k^m$ may have a different mean and covariance from $\boldsymbol{\nu}_k^m$. Thus, lines 10-11 estimate these as $\overline{\mathbf{u}}_{i,k}$ and $\overline{\boldsymbol{\Sigma}}_k$. Lines 13-16 compute rollouts using $\overline{\boldsymbol{\nu}}_k^m$, while line 18 evaluates state and control costs. Line 19 calculates weights $w^m$ using $s^m$ from line 18 and the updated statistics.

In line 22, we update the control profile, but unlike MPPIwSGF, our update uses $\overline{\mathbf{u}}_{i, k}$ instead of $\mathbf{u}_{i, k}$. Also, the noise samples differ from the Gaussian ones in line 3, as they are updated based on $\overline{\boldsymbol{\nu}}_k^m$ (line 15). Finally, line 25 re-applies the projection filter to ensure feasibility after averaging in line 22.

\begin{figure*}
\centering
  \includegraphics[width=0.9\linewidth]{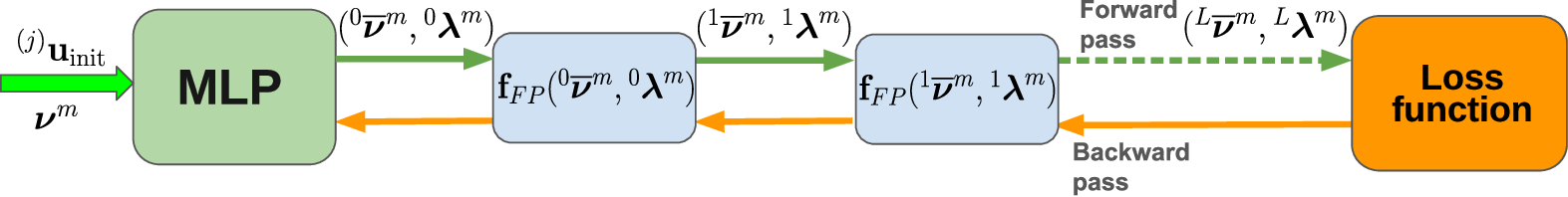} 
\caption{\footnotesize Our warm-start policy network. It consists of an MLP network and an unrolled chain of $L$ iterations of $\mathbf{f}_{FP}$. During training, the gradients flow through $\mathbf{f}_{FP}$, which ensures that the warm-start produced by MLP is aware of how it is going to be used by downstream $\mathbf{f}_{FP}$ ( steps \eqref{eqn:projection_cost}-\eqref{eqn:projection_lambda}  )  }
\label{fig:NN_block_diagram}
\vspace{-0.6cm}
\end{figure*} 

\subsection{Projection Filter}
\noindent Our projection filter is an optimization problem with the following form.
\begin{subequations}
\begin{align}
    &\min_{\overline{\boldsymbol{\nu}}^m} \frac{1}{2}\left\lVert  \boldsymbol{\nu}^m-\overline{\boldsymbol{\nu}}^m \right\rVert_2^2, \label{eqn:proj_nu_cost}\\
    &\text{s.t. } ^{(j)}\overline{\boldsymbol{\nu}}_0^m = \ ^{(j)}\mathbf{u}_{\mathrm{init}},\label{eqn:proj_eq_cons}\\
    &\quad \  ^{(j)}\mathbf{u}_{\mathrm{min}} \leq \,^{(j)}\overline{\boldsymbol{\nu}}^m \leq \,^{(j)}\mathbf{u}_{\mathrm{max}}. \label{eqn:proj_ineq_cons}
\end{align}    
\end{subequations}

\noindent As shown in Alg.~\ref{alg:one} (line 6), $\boldsymbol{\nu}^m$ is formed by stacking the respective values at different time-step $k$. The optimization \eqref{eqn:proj_nu_cost}-\eqref{eqn:proj_ineq_cons} minimally changes the $m^{th}$ perturbed control sequence to satisfy the initial conditions \eqref{eqn:proj_eq_cons} and bound \eqref{eqn:proj_ineq_cons} its $j^{th}$ derivatives. Eqns. \eqref{eqn:proj_nu_cost}-\eqref{eqn:proj_ineq_cons} define a QP. For example, if we approximate the control derivatives through finite-differencing, we  re-write \eqref{eqn:proj_nu_cost}-\eqref{eqn:proj_ineq_cons} into the following more compact form as:
\begin{subequations}
\begin{align}
    &\min_{\overline{\boldsymbol{\nu}}^m, \boldsymbol{\zeta}} \frac{1}{2}\left\lVert  \boldsymbol{\nu}^{m}-\overline{\boldsymbol{\nu}}^{m} \right\rVert_2^2, \label{eqn:proj_nu_cost_qp}\\
    &\text{s.t. } \mathbf{A}\overline{\boldsymbol{\nu}}^m = \mathbf{b},\label{eqn:proj_eq_cons_qp}\\
    &\quad \  \mathbf{G}\overline{\boldsymbol{\nu}}^m -\mathbf{h}-\boldsymbol{\zeta}= \mathbf{0}, \label{eqn:proj_ineq_cons_qp}\\
   &\quad \ \boldsymbol{\zeta}\geq \mathbf{0}, \label{slack}
\end{align}    
\end{subequations}

\noindent where $\overline{\boldsymbol{\nu}}^m$ is formed by stacking $\overline{\boldsymbol{\nu}}_k^m$ at different time-step $k$. A similar construction is followed for $\boldsymbol{\nu}^m$. The constant matrices $\mathbf{A},\mathbf{G}$ are formed with finite-difference matrices of up to $j^{th}$ order. Eqn.~\eqref{eqn:proj_eq_cons_qp} is the matrix-vector representation of \eqref{eqn:proj_eq_cons}, wherein the vector $\mathbf{b}$ is formed by stacking the initial conditions of the control sequence and their derivatives. Similarly, \eqref{eqn:proj_ineq_cons_qp} is derived from \eqref{eqn:proj_ineq_cons} by introducing a slack variable $\boldsymbol{\zeta}\geq 0$. Here, the vector $\mathbf{h}$ is simply formed by appropriately repeating and stacking ${^{(j)}}\mathbf{u}_{\mathrm{min}}$ and ${^{(j)}}\mathbf{u}_{\mathrm{max}}$. 



\subsubsection{Custom QP solver} In principle, any QP solver can be used for \eqref{eqn:proj_nu_cost_qp}-\eqref{eqn:proj_eq_cons_qp}. However, in order to ensure computational efficiency, we need to ensure that the following features are included in the solver: First, it should be batchable. That is, the solver should allow processing \eqref{eqn:proj_nu_cost_qp}-\eqref{eqn:proj_eq_cons_qp}, $\forall m$ in parallel. Moreover, it would be advantageous if this parallelization could leverage GPU acceleration. Second, it should be possible to learn good warm-start for the employed QP solver. As we discuss later, this, in turn, requires that every step of the QP solver should consist of only differentiable operations.

We fulfill both the above-mentioned requirements by adapting the Alternating Direction Method of Multipliers (ADMM) based QP solver from \cite{ghadimi2014optimal}. First, we relax the equality constraints \eqref{eqn:proj_ineq_cons_qp} as Euclidean norm penalties and append them into the cost function to construct the following augmented Lagrangian.
\begin{dmath}
    \mathcal{L} (\overline{\boldsymbol{\nu}}^m, \boldsymbol{\lambda}^m, \boldsymbol{\zeta}^m) =  \frac{1}{2}
    \left\lVert \boldsymbol{\nu}^m-\overline{\boldsymbol{\nu}}^m \right\rVert_2^2- \langle \boldsymbol{\lambda}^m,\overline{\boldsymbol{\nu}}^m\rangle +\frac{\delta}{2}\left\lVert \mathbf{G}\overline{\boldsymbol{\nu}}^m-\mathbf{h}+\boldsymbol{\zeta}^m  \right\rVert_2^2,
    \label{aug_lag}
\end{dmath}

\noindent where $\boldsymbol{\lambda}^m$ is the so-called Lagrange multiplier and  $\delta$ is the weight of constraint penalty. We minimize \eqref{aug_lag} subject to \eqref{eqn:proj_eq_cons_qp} and \eqref{slack} through the iterative steps \eqref{eqn:projection_cost}-\eqref{eqn:projection_lambda}, where the left super-script signifies the value of the variable at iteration $l$. For example, $^{l}\boldsymbol{\lambda}^m$ is the value of $\boldsymbol{\lambda}^m$ at iteration $l$.
\begin{subequations}
\begin{align}
    &\,^{l+1}\boldsymbol{\zeta}^m = \max \left( 0, -\mathbf{G}\, ^{l}\overline{\boldsymbol{\nu}}^m-\mathbf{h}\right),\label{eqn:projection_slack}\\
    &^{l+1}\overline{\boldsymbol{\nu}}^m =  \arg \min_{\overline{\boldsymbol{\nu}}^m} \mathcal{L}(\overline{\boldsymbol{\nu}}^m, {^l}\boldsymbol{\lambda}^m, \,^{l+1}\boldsymbol{\zeta}^m), \,\, \mathbf{A}\overline{\boldsymbol{\nu}}^m = \mathbf{b} \label{eqn:projection_cost}\\
    & \qquad \quad  = \mathbf{D}^{-1}\mathbf{d}(\boldsymbol{\nu}^m,\,^{l}\boldsymbol{\lambda}^m,\,^{l+1}\boldsymbol{\zeta}^m),\\
   &\,^{l+1}\boldsymbol{\lambda}^m = \,^{l}\boldsymbol{\lambda}^m - \frac{\delta}{2}\mathbf{G}^{\intercal}(\mathbf{G}\, ^{l+1}\overline{\boldsymbol{\nu}}^m-\mathbf{h}+\,^{l+1}\boldsymbol{\zeta}^m),\label{eqn:projection_lambda}
\end{align}    
\end{subequations}
\small
\begin{align}
    \mathbf{D}=\begin{bmatrix}
    \mathbf{I}+ \delta\mathbf{G}^{\intercal}\mathbf{G} & \mathbf{A}^{\intercal} \\
    \mathbf{A} & \mathbf{0} 
\end{bmatrix}, \mathbf{d}=\begin{bmatrix}
    \delta\mathbf{G}^{\intercal}(\mathbf{h}+\,^{l+1}\boldsymbol{\zeta}^m)+\,^{l}\boldsymbol{\lambda}^m+\boldsymbol{\nu}^m \\ \mathbf{b}
\end{bmatrix} 
\end{align}
\vspace{-0.7cm}
\normalsize

\noindent At iteration $l+1$, we first compute $^{l+1}\boldsymbol{\zeta}^m$ using ${^{l}}\overline{\boldsymbol{\nu}}^m$. Importantly, this step does not involve any matrix factorization. Then we minimize $\mathcal{L}$ while fixing $\boldsymbol{\lambda}^m$ at values obtained in the previous iteration and $\boldsymbol{\zeta}^m$ obtained from \eqref{eqn:projection_slack}. As clear from \eqref{eqn:projection_cost}, this minimization is essentially an equality-constrained QP with a closed-form solution. Step \eqref{eqn:projection_lambda} updates the Lagrange multipliers based on ${^{l+1}}\boldsymbol{\overline{{\nu}}}^m$ and ${^{l+1}}\boldsymbol{\zeta}^m$.

\noindent \emph{Batch Computation Over GPUs:} Steps \eqref{eqn:projection_slack}-\eqref{eqn:projection_lambda} just involves matrix-vector product that can be easily parallelized over GPUs. The step \eqref{eqn:projection_cost} involves a matrix factorization. But interestingly, the matrix $\mathbf{D}$ is independent of the input $\boldsymbol{\nu}^m$ or the projection variable $\overline{\boldsymbol{\nu}}^m$. Thus, it remains the same for all the $m$ projection problems in line 7 of Alg.~\ref{alg:one}. Moreover, the factorization needs to be done only once and can be pre-stored before the start of our $\pi-$MPPI pipeline employed for MPC.

\subsection{Learning to warm start the QP solver}\label{sec:learning}
\noindent The steps \eqref{eqn:projection_cost}-\eqref{eqn:projection_lambda} are guaranteed to converge to the optimal solution of \eqref{eqn:proj_nu_cost_qp}-\eqref{slack} \cite{ghadimi2014optimal}. However, the quality of initialization can affect the convergence speed. Thus, the objective of this sub-section is to learn a neural network policy that maps in observations to initialization $({^0}\overline{\boldsymbol{\nu}}^m, {^0}\boldsymbol{\lambda}^m)$ for the QP solver's steps \eqref{eqn:projection_cost}-\eqref{eqn:projection_lambda}. The observations consist of the unfiltered/un-projected perturbed control sequence and their boundary conditions, 
$(\boldsymbol{\nu}^m, {^{(j)}}\mathbf{u}_{\mathrm{init}})$. The learned initialization will accelerate the convergence of the \eqref{eqn:projection_cost}-\eqref{eqn:projection_lambda}. To this end, we build on two key insights. First, the steps \eqref{eqn:projection_cost}-\eqref{eqn:projection_lambda} can be viewed as a fixed point iteration of the following form:
\begin{align}
    ({^{l+1}}\overline{\boldsymbol{\nu}}^m, {^{l+1}}\boldsymbol{\lambda}^m) = \mathbf{f}_{FP} ({^{l}}\overline{\boldsymbol{\nu}}^m, {^{l}}\boldsymbol{\lambda}^m)
\end{align}

\noindent Second, $\mathbf{f}_{FP}$ is differentiable as \eqref{eqn:projection_cost}-\eqref{eqn:projection_lambda} only involves differentiable operations. The differentiable fixed-point interpretation allows us to build on \cite{sambharya2024learning} and design the learning pipeline shown in Fig.~\ref{fig:NN_block_diagram}. It consists of a multi-layer perceptron (MLP) network and unrolled chain of $\mathbf{f}_{FP}$ iterations. The MLP takes in the observation $(\boldsymbol{\nu}^m, {^{(j)}}\mathbf{u}_{\mathrm{init}})$
and produces the initialization $({^0}\overline{\boldsymbol{\nu}}^m, {^0}\boldsymbol{\lambda}^m)$. These are then passed through $L$ iterations of $\mathbf{f}_{FP}$ (i.e \eqref{eqn:projection_cost}-\eqref{eqn:projection_lambda}). The learnable parameters are contained only in the MLP part, and they are trained with the following loss function:
\begin{align}
    \min_{\boldsymbol{\Theta}} \sum_l \left \Vert \begin{bmatrix}
        {^{l+1}}\overline{\boldsymbol{\nu}}^m\\
        {^{l+1}}\boldsymbol{\lambda}^m
    \end{bmatrix}-\mathbf{f}_{FP} ({^{l}}\overline{\boldsymbol{\nu}}^m, {^{l}}\boldsymbol{\lambda}^m) \right\Vert_2^2+\left\Vert {^L}\overline{\boldsymbol{\nu}}^m-\boldsymbol{\nu}^m\right\Vert_2^2,
    \label{NN_loss}
\end{align}
\noindent where $\boldsymbol{\Theta}$ are the learnable weights of the MLP. The first term in the loss function \eqref{NN_loss} minimizes the fixed-point residual at each iteration, which in turn accelerates the convergence of the iterative projection \eqref{eqn:projection_cost}-\eqref{eqn:projection_lambda}. The second term ensures that MLP produces initialization, which leads to the correction of the infeasible control sequence $\boldsymbol{\nu}^m$ through minimum possible change.

Some important features of our learning pipeline are worth pointing out. First, the training is self-supervised. That is, we don't need a dataset that stores optimal solution $\overline{\boldsymbol{\nu}}^m$ corresponding to the input $\boldsymbol{\nu}^m$ for a large number of cases. Second, during training, the gradients of the loss function are traced through the iterations of $\mathbf{f}_{FP}$, which in turn necessitates its differentiability. Moreover, this also ensures that the MLP is aware of how its generated initialization is used in the downstream $\mathbf{f}_{FP}$ operations. This dramatically improves the data efficiency of our learning pipeline. 


\subsection{Low Dimensional Control Parametrization}\label{sec:con_parametrization}
\noindent We now present an additional improvement that can be optionally integrated into the $\pi$-MPPI pipeline. We propose parameterizing each component of the control sequence as time-dependent polynomials similar to \cite{bhardwaj2022storm,pezzato2025sampling}.
\small
\begin{align}
    \begin{bmatrix}
        v_1, .., v_n 
    \end{bmatrix} = \mathbf{W}\mathbf{c}_{v},
    \begin{bmatrix}
        \phi_1, .., \phi_n 
    \end{bmatrix} = \mathbf{W}\mathbf{c}_{\phi},
     \begin{bmatrix}
        \theta_1, .., \theta_n 
    \end{bmatrix} = \mathbf{W}\mathbf{c}_{\theta},
    \label{param} 
\end{align}
\normalsize
\noindent where $\mathbf{W}$ is a constant matrix formed with time-dependent polynomial basis functions and $\mathbf{c}_v, \mathbf{c}_{\phi},  \mathbf{c}_{\theta} \in \mathbb{R}^{n}$ are coefficient vectors. We can write similar expressions for derivatives using $\dot{\mathbf{W}}, \ddot{\mathbf{W}}$. Using \eqref{param}, we can map the mean control trajectory $\mathbf{u}_{i}$ obtained at $i^{th}$ MPC iteration to an equivalent $(\mathbf{c}_{v, i}, \mathbf{c}_{\phi, i},  \mathbf{c}_{\theta, i})$.

One advantage of \eqref{param} is that it allows representing long-horizon control sequences through low dimensional coefficient vectors. That is, $n<<K$. We can modify the control perturbation step of MPPI in the following manner. Let $\boldsymbol{\epsilon}_v^m, \boldsymbol{\epsilon}_{\phi}^m, \boldsymbol{\epsilon}_{\theta}^m \in \mathbb{R}^{n}$ be noise samples drawn from some zero-mean Gaussian with covariance $\boldsymbol{\Sigma}_v, \boldsymbol{\Sigma}_{\phi}, \boldsymbol{\Sigma}_{\theta}$. Using \eqref{param}, we can rewrite control perturbation in the following manner.
\begin{align}
 \boldsymbol{\nu}_k^m = \mathbf{u}_{i, k} + \begin{bmatrix}
     \mathbf{W}_k & 0 & 0\\
     0 & \mathbf{W}_k & 0\\
     0 & 0 & \mathbf{W}_k\\
 \end{bmatrix}\begin{bmatrix}
     \boldsymbol{\epsilon}_{v, k}^m\\
     \boldsymbol{\epsilon}_{\phi, k}^m\\
     \boldsymbol{\epsilon}_{\theta, k}^m\\
 \end{bmatrix},
 \label{pert_control_matrix}
\end{align}

\noindent where $\mathbf{W}_k, \boldsymbol{\epsilon}_{v, k}^m, \boldsymbol{\epsilon}_{\phi, k}^m, \boldsymbol{\epsilon}_{\theta, k}^m$ represent the $k^{th}$ rows of the respective matrix and vectors.

The biggest advantage of polynomial control parameterization comes in the projection step. Assume that the components of the projected control sequence $\overline{\boldsymbol{\nu}}^m$ follow a similar parametrization as \eqref{param}. That is, we can parameterize $\overline{\boldsymbol{\nu}}^m$ through some vector $(\overline{\mathbf{c}}_v^m, \overline{\mathbf{c}}_{\phi}^m, \overline{\mathbf{c}}_{\theta}^m)$. Consequently, we can easily re-phrase the projection optimization \eqref{eqn:proj_nu_cost_qp}-\eqref{slack} in terms of  $(\overline{\mathbf{c}}_v^m, \overline{\mathbf{c}}_{\phi}^m, \overline{\mathbf{c}}_{\theta}^m )$. This dramatically reduces the dimensionality of the projection problem and improves computational efficiency. Along similar lines, we can restructure the warm-start learning pipeline to generate warm starts for the coefficients.

\section{Connections to Existing Works}
\noindent \textit{Connections to MPPI literature:} Our approach $\pi$-MPPI provides several benefits and interesting connections to the existing works. First, $\pi$-MPPI can be seen as adapting the input mean control sequence before computing the optimal update on it (line 10 Alg.~\ref{alg:one}). In this sense, it is connected to recent efforts such as \cite{tao2023rrt}, \cite{trevisan2024biased}. In a similar vein, $\pi$-MPPI can also be viewed from the point of adapting the noise distribution at each MPC iteration (line 11, Alg.~\ref{alg:one}), but without using concepts from evolutionary optimization proposed in \cite{bhardwaj2022storm}, \cite{asmar2023model} or learning \cite{sacks2023learning}, \cite{power2022variational}. Finally, in baseline MPPIwSGF \cite{williams2017model}, \cite{williams2018information}, the covariance matrix of the Gaussian distribution from where the control perturbation noise is sampled (line 3 in Alg.~\ref{alg:one}) needs to be carefully crafted. A large covariance can lead to highly noisy trajectories and consequently diverging of the optimal control update rule \eqref{eqn:mean_u} \cite{mohamed2022autonomous}. In contrast, $\pi$-MPPI can allow for arbitrary covariance as the projection operation is guaranteed to ensure smooth control profile and state trajectory rollouts. This feature, in turn, can be leveraged to increase the trace of the covariance matrix in $\pi$-MPPI and induce more efficient exploration.  

Our work addresses the same problem as \cite{kim2022smooth}, but in a more generalizable manner. Specifically,  \cite{kim2022smooth}  samples $\dot{\boldsymbol{\nu}}_k^m$ and treat $\boldsymbol{\nu}_k^m$ as one of the state components. They also augment the cost function with penalties on $\boldsymbol{\nu}_k^m$. Such a modification introduces the additional challenge of tuning the cost function to appropriately trade off the primary cost and the penalties on $\boldsymbol{\nu}_k^m$. This tuning challenge can be further aggravated if we want to introduce bounds on $j^{th}$ order derivatives of $\boldsymbol{\nu}_k^m$, as this would require increasing the system dynamics by $j$ and introducing $j-1$ control penalties in the cost function. In contrast, our work can induce an arbitrary degree of smoothness and does not introduce any additional penalties in the cost function.

\noindent \textit{Learning to Warm-Start Literature:} A good initialization can significantly speed up the convergence of optimization solvers. Early efforts focused on using supervised learning on optimal trajectory datasets to initialize solvers \cite{mansard2018using}, \cite{eiras2021two}. However, recent work shows that a ``good initialization" is specific to the solver's numerical algebra \cite{sambharya2024learning}. In other words, learned initialization is less effective if the model doesn't consider how its predictions will be used by the solver. Our prior work \cite{rastgar2024priest} also found that warm-starting solvers with solutions from a global optimizer yielded minimal benefits. Therefore, \cite{sambharya2024learning} suggests embedding the solver within the learned model, as shown in Fig.~\ref{fig:NN_block_diagram}, which requires differentiable solver steps, achieved here with \eqref{eqn:projection_cost}-\eqref{eqn:projection_lambda}.

\section{Validation and Benchmarking}\label{sec:results}

\begin{figure}[!t]
\begin{subfigure}{4.2cm}
  \includegraphics[width=\linewidth,height=2.5cm]{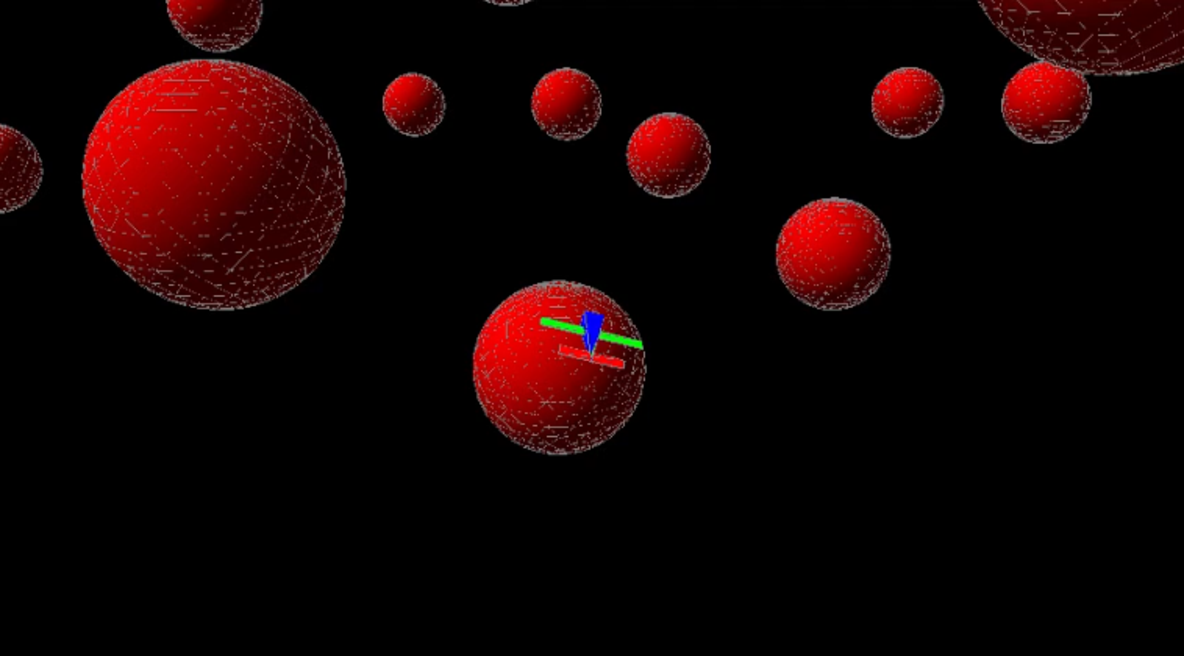}  
  \caption{}
  \label{fig:obs_scene}
\end{subfigure}
\begin{subfigure}{4.2cm}
  \includegraphics[width=\linewidth,height=2.5cm]{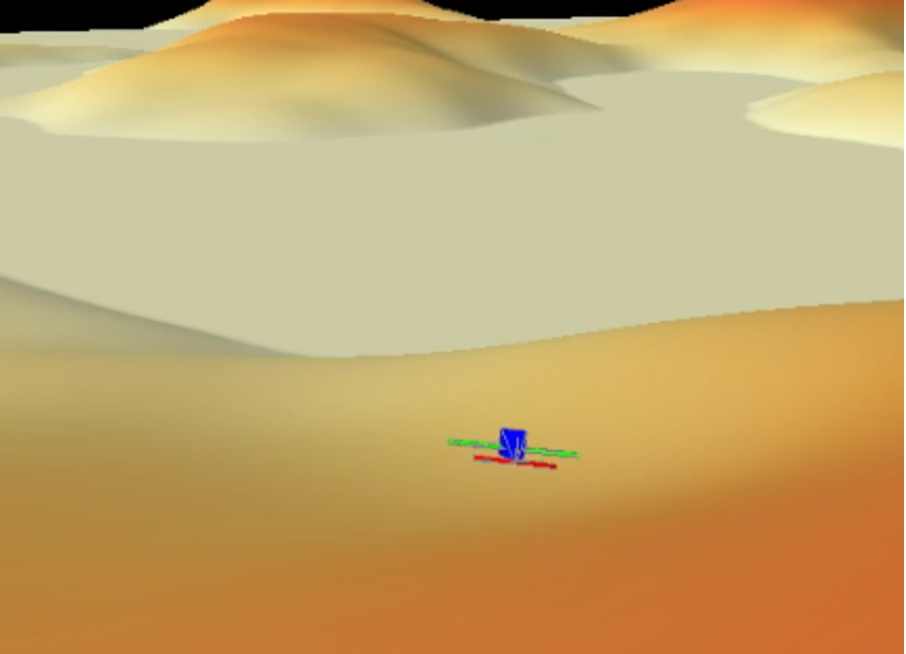}  
  \caption{}
  \label{fig:terrain_scene}
\end{subfigure}
\newline
\begin{subfigure}{4.2cm}
  \includegraphics[width=\linewidth,height=3cm]{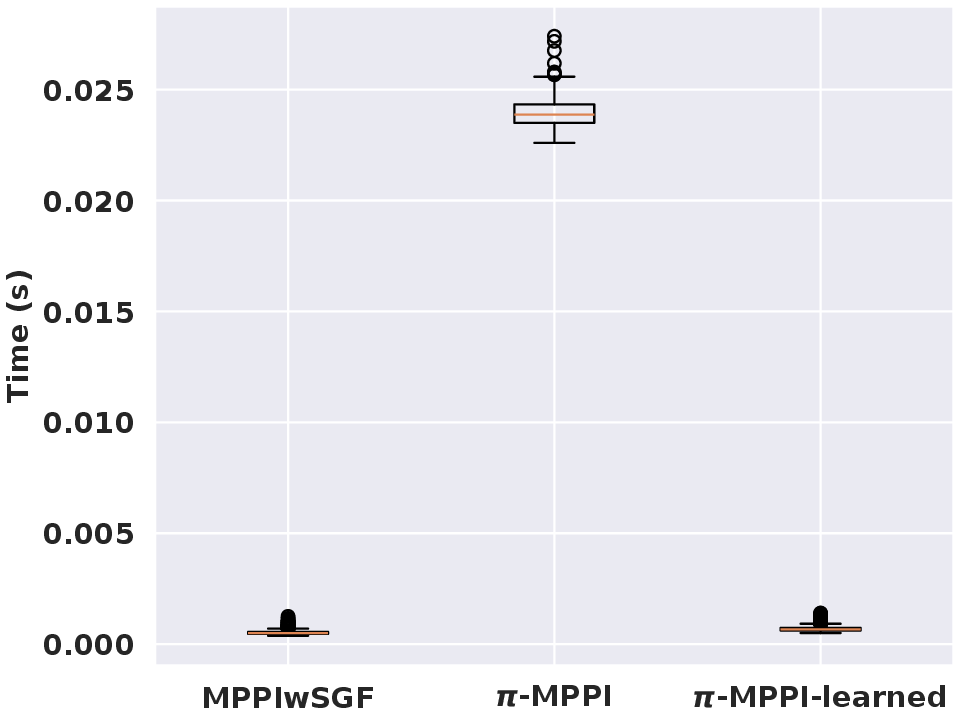}  
  \caption{}
  \label{fig:comp_time}
\end{subfigure}
\begin{subfigure}{4.2cm}
  \includegraphics[width=\linewidth,height=3cm]{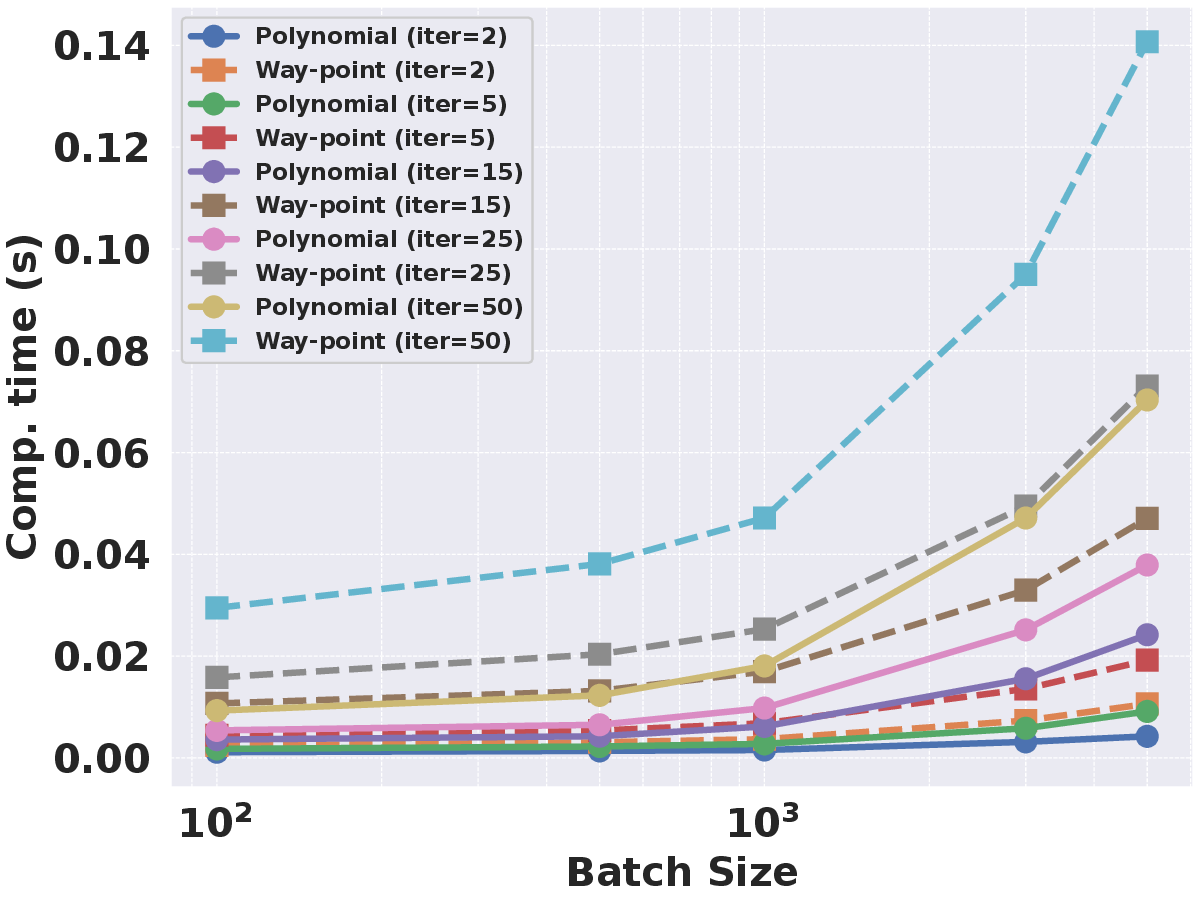} 
\caption{}
\label{fig:poly_time_comparison}
\end{subfigure} 
\vspace{-0.3cm}
\caption{\footnotesize Benchmarks used to evaluate $\pi$-MPPI against baselines and comparison of computation-time. a) Obstacle avoidance scenario b) \footnotesize Terrain following scenario c) \footnotesize Computation time per MPC step. d) Computation time for solving \eqref{eqn:proj_nu_cost_qp}-\eqref{slack} with polynomial and way-point parametrization of $\overline{\boldsymbol{\nu}}^m$, for varying batch size and number of projection iterations.}
\label{fig:scene}
\vspace{-0.6cm}
\end{figure}


This section aims to (1) assess $\pi$-MPPI's ability to generate smooth, bounded control sequences compared to existing MPPI variants, (2) evaluate its impact on task performance, and (3) examine whether learned warm-start improves its computational efficiency to match MPPIwSGF. Most qualitative results are in the accompanying video due to space constraints.



\subsection{Implementation Details}
\noindent Alg. \ref{alg:one} was implemented in Python using the JAX library for GPU-accelerated computing. The warm-start policy was trained using the Pytorch library. The tunable parameters in the MPPI algorithm are selected as $\sigma=5$ and $\alpha=0.99$. We used planning horizon $K=100$ with a sampling time of $0.2$\,s. We computed $M=1000$ rollouts of the dynamics. For $\pi$-MPPI, we used the polynomial parameterization of the control sequence, introduced in \ref{sec:con_parametrization}. Without neural warm-start, we ran the projection filter for 50 iterations. The bounds on the control input and derivatives were chosen by observing the response of an FWV for different control trajectories in a flight simulator.

The MLP used in Fig.~\ref{fig:NN_block_diagram} as a part of the warm-start policy for the projection filter (QP solver) had two hidden layers of size 1024 and 256, respectively. We used ReLU activation in the hidden layers. The output layer had 66 neurons and used a linear activation function. The input layer mapped the obervations (${^{(j)}}\mathbf{u}_{\mathrm{init}}, \boldsymbol{\nu}^m)$ to a 1024 dimensional vector. To train our warm-start policy network, we ran $\pi$-MPPI with zero initialization of the projection filter to create a dataset of different (${^{(j)}}\mathbf{u}_{\mathrm{init}}, \boldsymbol{\nu}^m)$.

\subsubsection{Baselines}
\noindent We use $\pi$-MPPI in an MPC setting and compare it with the following baselines. Note that all the baselines are given 4 times higher control samples $(M=4000)$ than $\pi$-MPPI. Moreover, we did an extensive hyperparameter search for the temperature $\sigma$, and the best results were reported.\\
\noindent \textbf{MPPIwSGF:} This is the standard MPPI pipeline from \cite{williams2018information}. Comparing MPPIwSGF and $\pi$-MPPI is tricky since the former imposes fewer constraints on the resulting motions. Thus, we adopt the following approach. We tune the control covariance of MPPIwSGF so that the violations in the control derivatives are within some respectable limits.  \\
\noindent \textbf{MPPIwSGF-high-cov:} This variant is obtained by using larger control covariances compared to the baseline MPPIwSGF. The objective is to evaluate if extra noise-induced exploration can improve task completion.\\
\noindent \textbf{MPPIwSGF-E:} This is an extension of baseline MPPIwSGF. We follow \cite{wagener2019online}, which shows that the control-update \eqref{eqn:mean_u} can also be derived from the point of view of online learning. Interestingly, this allows for arbitrary control costs. We leverage this insight and incorporate additional terms in the cost function that model the violation of bounds in control and their derivatives. More specifically, for a constraint of the form $g(\xi)\leq 0$, we can add penalties of the form $\max(0, g(\xi))$ into the cost function. However, this baseline requires additional effort to appropriately tune the weights of the primary cost and control penalties\\
\noindent \textbf{MPPI-poly:} This is a variant with polynomial parametrization for the noise samples $\boldsymbol{\epsilon}_{v}$,$\boldsymbol{\epsilon}_{\phi}$, and $\boldsymbol{\epsilon}_{\theta}$. (see \eqref{param}). Here again, we tune the control covariance to balance control derivatives violation and task performance. 
\subsubsection{Benchmarks and Tasks}
\noindent We consider the following two benchmarks that are also shown in Fig.~\ref{fig:scene}. We performed 250 experiments for each benchmark, and each one of them involved 1000 MPC steps. 
\\
\noindent \textit{Static Goal Tracking With Obstacle Avoidance:} In this benchmark (Fig.~\ref{fig:obs_scene}), the FWV is required to converge to a static goal and continuously encircle it while avoiding a set of 3D obstacles. We consider a 3D space with dimensions 2000x2000x400\,m filled with 100 obstacles with radii ranging from 20\,m to 30\,m. The covariance for the MPPIwSGF is $\boldsymbol{\Sigma}_k=diag([0.02,0.0002,0.0003])$, for MPPIwSGF-high-cov is $\boldsymbol{\Sigma}_k=diag([0.1,0.001,0.0015])$, and for MPPI-poly is $\boldsymbol{\Sigma}_k=diag([0.02,0.0002,0.002])$. In $\pi$-MPPI, we used the polynomial parameterization for the control sequence and injected noise into the polynomial coefficients, which are mapped back to control using \eqref{pert_control}. The covariance matrix for sampling $\boldsymbol{\epsilon}_{v}$,$\boldsymbol{\epsilon}_{\phi}$, and $\boldsymbol{\epsilon}_{\theta}$ was $\boldsymbol{\Sigma}_v = 20\mathbf{I}$, $\boldsymbol{\Sigma}_{\phi} = 2\mathbf{I}$, and $\boldsymbol{\Sigma}_{\theta}= 2\mathbf{I}$ respectively, wherein $\mathbf{I}$ is an identity matrix.

The primary cost $c(\cdot)$ comprises a goal reaching cost $\left\lVert \mathbf{p}_k-\mathbf{p}_g \right\rVert_2^2$ and an obstacle avoidance cost, $\sum_o \max\left(0,-\left(\left\lVert \mathbf{p}_k-\mathbf{p}_o \right\rVert_2 - (r_{o} + r_{fwv})\right)\right)$, where $\mathbf{p}_k=[p_{n,k},p_{e, k},p_{d, k}]$ is the position of the FWV. The vector $\mathbf{p}_g$ is the goal point, $\mathbf{p}_o$ is the position of the $o$\textsuperscript{th} obstacle with radius $r_{o}$, and $r_{fwv}$ is the radius of an imaginary sphere enclosing the FWV. \\ 
\noindent \textit{Static Goal Tracking with Terrain Following:} In this benchmark, the task is to track a static goal while the FWV stays within the given minimum and maximum distance to the terrain surface. We consider an uneven terrain with dimensions 4000x4000x100\,m. The covariance for the MPPIwSGF is $\boldsymbol{\Sigma}_k=diag([0.008,0.0015,0.006])$, for MPPIwSGF-high-cov is $\boldsymbol{\Sigma}_k=diag([0.04,0.0075,0.03])$, and for MPPI-poly is $\boldsymbol{\Sigma}_k=diag([0.02,0.0002,0.002])$. The $\pi$-MPPI with polynomial control parameterization used $\boldsymbol{\Sigma}_v = 100\mathbf{I}$, $\boldsymbol{\Sigma}_{\phi} = 3\mathbf{I}$, $\boldsymbol{\Sigma}_{\theta} = 0.6\mathbf{I}$ to sample noise for the coefficients (recall \eqref{pert_control}).

The primary cost $c(\cdot)$ comprises a goal reaching cost $\sqrt{(p_{n,k}-p_{ng})^2+(p_{e,k}-p_{eg})^2}$ and a terrain following cost, $\max(0,-(z_{fwv,k}-z_{min}))$+$\max(0,(z_{fwv,k}-z_{max}))$, where $(p_{ng},p_{eg})$ is the 2D goal point, $z_{fwv,k}=-p_{d,k}$ is the altitude of the FWV and ($z_{min}$, $z_{max})$ are the minimum and maximum allowed altitudes. $z_{min}$ is selected according to the size of the FWV, and $z_{max}$ dictates how close the FWV follows the terrain. For the simulations, $z_{min}$ is selected as 5\,m above the terrain surface and $z_{max}$ as 15\,m above the surface. Note, as shown in Fig.~\ref{fig:scene}, the terrain surface has a highly complex shape with varying heights. Thus, the altitude of the terrain surface and, consequently, the actual numerical value of $z_{min}$ and $z_{max}$ depends on the 2D position of the FWV.

\subsubsection{Metrics} We consider the following three metrics in our analysis.

\noindent \textit{Success-rate:} Number of simulation experiments out of total in which the FWV does not collide with the obstacle (first benchmark) or terrain (second benchmark).\\
\noindent \textit{Commanded Control Smoothness:} Let $\mathbf{u}_{i,0}$ be the commanded control input at the $i^{th}$ MPC iteration. We can define the rate and double rate of change of the commanded acceleration as $\dot{\mathbf{u}}_{i,0} = \frac{\mathbf{u}_{i+1,0}-\mathbf{u}_{i,0}}{\Delta t_{sim}}$, $\ddot{\mathbf{u}}_{i,0} = \frac{\dot{\mathbf{u}}_{i+1,0}-\dot{\mathbf{u}}_{i,0}}{\Delta t_{sim}}$, where $\Delta t_{sim}$ is the simulation update step. We use constraint residuals as one of the metrics that quantify how much $\mathbf{u}_{i,0}$, $\dot{\mathbf{u}}_{i,0}$, and $\ddot{\mathbf{u}}_{i,0}$ violate their respective bounds. For constraints of the form $g(\xi)\leq 0$, the residuals are simply $\max(0, g(\xi))$. This metric is designed to showcase that a higher degree of smoothness in the planned trajectory leads to smoothness across the commanded control inputs at each MPC iteration. We recall that $\mathbf{u}_{i,0}$ will have three components $v_{i,0}, \phi_{i,0}, \theta_{i,0}$. The derivatives will have the same number of components.   \\
\noindent \textit{Average distance-to-the-goal:} This metric captures the average distance at which FWV is able to track/en-circle the static goal. 
\vspace{-0.2cm}
\subsection{Qualitative and Quantitative Comparison}

\begin{figure*}[!h]
    \centering
    \includegraphics[scale = 0.3]{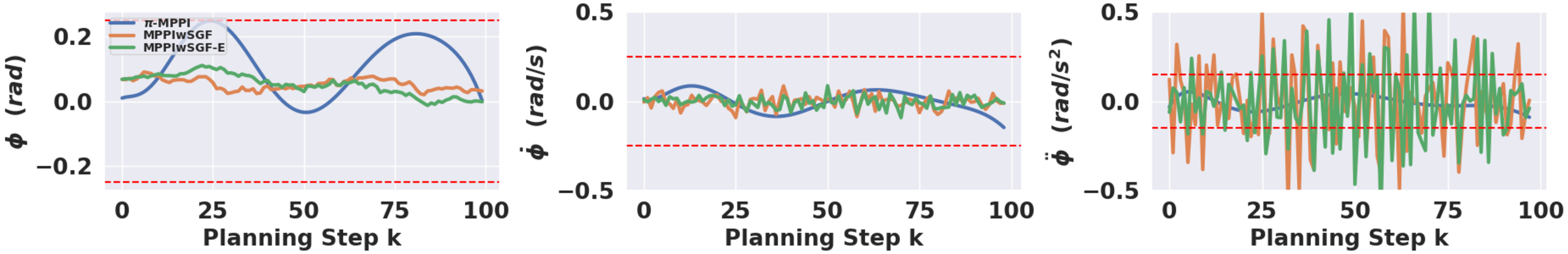}
    \caption{\footnotesize{Plot of the optimal roll angle and its derivatives resulting from $\pi$-MPPI and baselines.}}
    \label{fig:smooth_controls_plan}
    \vspace{-0.4cm}
\end{figure*}

\begin{figure*}[!h]
    \centering
    \includegraphics[scale = 0.32]{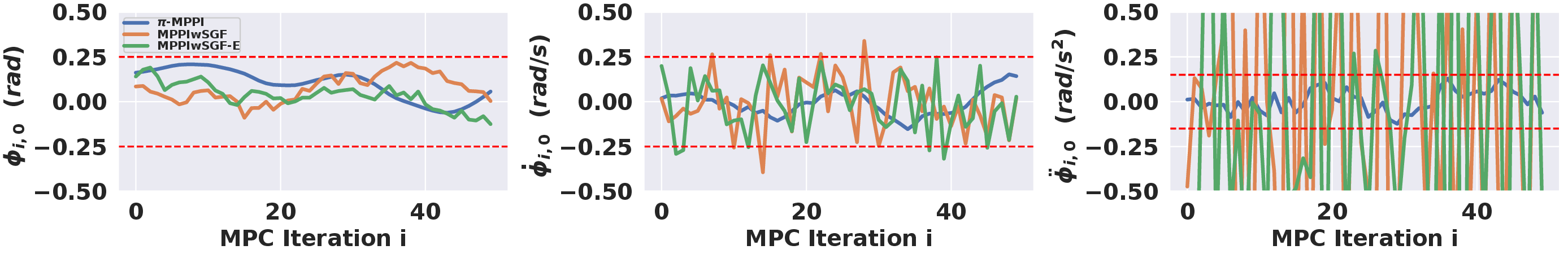}
    \caption{\footnotesize{Plot of the commanded roll angle across the first 50 MPC iteration for one of the experiments. It can be seen that our approach $\pi$-MPPI produces the smoothest profile that satisfies bounds on not only control values but also its first and second derivatives.}}
    \label{fig:smooth_controls}
    \vspace{-0.6cm}
\end{figure*}

\noindent Fig.~\ref{fig:smooth_controls_plan} shows the planned control sequence from $\pi$-MPPI and baselines for an example case from the terrain following scenario, while Fig.~\ref{fig:smooth_controls} shows the plot of commanded control across the MPC iteration $i$. For lack of space, we only show the planned ($\boldsymbol{\phi}$) and commanded roll angle ($\phi_{i,0}$). It can be seen that {MPPIwSGF} can only satisfy the bounds on just the control values while the constraints on the derivatives are violated. The baseline {MPPIwSGF-E} is able to leverage its control penalties to suppress some of these violations. However, $\pi$-MPPI produces the smoothest planned and commanded roll angle profile that satisfies both first and second-order bounds.

Table \ref{table:obs_and_terrain_base_vs_proj} further quantifies the smoothness gains provided by $\pi$-MPPI for the obstacle avoidance and terrain following benchmark. It can be seen that  MPPIwSGF has lower constraint residuals than MPPIwSGF-high-cov. MPPIwSGF-E is able to leverage the control penalties to drive the residuals lower than MPPIwSGF. MPPI-poly has slightly lower residuals than MPPIwSGF due to the smoothness introduced by the polynomial parametrization. Our approach $\pi$-MPPI outperforms all the baselines by producing the lowest residuals of constraints on control and its derivatives.


The success-rate trend also sheds interesting insights into the baseline. Due to its larger exploration, MPPIwSGF-high-cov has a higher success rate in both benchmarks than MPPIwSGF. MPPI-Poly performance is somewhere between MPPIwSGF and MPPIwSGF-high-cov. MPPIwSGF-E's performance is very erratic: it did well in the terrain following scenario but not in the obstacle avoidance benchmark. This exhibits the difficulty in balancing the effects of control penalties and other state costs. Our approach $\pi$-MPPI provides a noticeable improvement over all the baselines. This is due to two key reasons:
\begin{itemize}
    \item For the baselines, the inherent non-smoothness in the control profile means that the input sequence $\boldsymbol{u}_{i,k}$ between two MPC iterations can oscillate sharply (see Fig.~\ref{fig:smooth_controls}. This is particularly true when the control covariance is high. This, in turn, can disturb the warm-starting of the MPPI pipeline, which is critical for its success. In contrast, $\pi$-MPPI ensures that $\boldsymbol{u}_{i,k}$ between two consecutive MPC steps are close due to the bounds on the control derivatives.
    \item $\pi$-MPPI can withstand large control perturbation noise and, thus, create rollouts that explore the state space better (see accompanying video). Consequently, we achieve lower values for state-space costs such as collision. In contrast, increasing the covariance of baselines is very tricky as it can further accentuate non-smoothness in the control (Table~\ref{table:obs_and_terrain_base_vs_proj}) and aggravate the problem discussed in the previous point.
\end{itemize}
\vspace{-0.22cm}



\subsection{Effect of Warm-starting}

\noindent This section demonstrates how our learned warm-start policy reduces projection optimization iterations with minimal impact on task metrics. We denote it as $\pi$-MPPI-learned, distinguishing it from $\pi$-MPPI, which uses zero initialization. We compare constraint residuals on $(\mathbf{v}, \boldsymbol{\phi}, \boldsymbol{\theta})$ and their derivatives for 2 and 5 projection iterations, averaging per-time step residuals over MPC experiments.

Table \ref{tab:leanred_vs_proj_comparison} summarizes the benchmark results. In obstacle avoidance, $\pi$-MPPI-learned yields lower residuals for most constraints, except velocity and its derivatives, with fewer iterations. It also achieves a higher success rate and shorter average distance to the goal with a reduced iteration budget.  

A similar trend is observed in the terrain following. The warm-start neural policy in $\pi$-MPPI-learned significantly improves constraints on $\mathbf{v}, \boldsymbol{\phi}, \boldsymbol{\dot{\phi}}, \boldsymbol{\ddot{\phi}}$, while both methods achieve minimal residuals for other components. This leads to a higher success rate and shorter distance to the goal for $\pi$-MPPI-learned.

\vspace{-0.2cm}
\subsection{Ablations}
\noindent We analyze alternate methods to warm-start the QP solver \eqref{eqn:projection_slack}-\eqref{eqn:projection_lambda} besides our neural policy shown in Fig.~\ref{fig:NN_block_diagram} (used in $\pi$-MPPI-learned). The first ablation, $\boldsymbol{\nu}^m$-initialization, initializes the projection optimizer with perturbed control samples $\boldsymbol{\nu}^m$ instead of zeros or neural network values. The second, `Direct Solution,’ trains a neural network to predict the projection solution directly. Table \ref{tab:learned_initialization_comparison} compares these methods on the benchmark scenarios, showing that the neural warm-start policy achieves the lowest constraint residuals after 2 projection iterations. Comparisons at other iteration limits are presented in the accompanying video.
\vspace{-0.33cm}

\begin{table*}[htp]
\centering
\scriptsize

\begin{NiceTabular}{*{1}{p{1.4cm}}*{1}{p{1cm}}*{1}{p{1cm}}*{1}{p{1cm}}*{1}{p{1cm}}*{1}{p{1cm}}*{1}{p{1cm}}*{1}{p{1cm}}*{1}{p{1cm}}*{1}{p{1cm}}*{1}{p{1cm}}*{1}{p{1cm}}}[hvlines]

\Block{2-2}{Metrics} & & \multicolumn{5}{c}{Obstacle avoidance scenario} & \multicolumn{5}{c}{Terrain following scenario} \\
 & & MPPI wSGF & MPPI wSGF-high-cov & MPPI wSGF-E & MPPI-poly & $\pi$-MPPI & MPPI wSGF & MPPI wSGF-high-cov & MPPI wSGF-E & MPPI-poly & $\pi$-MPPI \\
\Block{1-2}{Success rate (\%)} & & 90.4 & 93.6 & 78.4 & 94.0 & \textbf{99.6} & 92 & 97.2 & 96.8 & 91.6 & \textbf{99.2} \\
\Block{1-2}{Avg. dist. to goal (m)} & & 87.786 & 81.173 & 92.261 & 81.211 & 103.960 & 235.252 & 248.234 & 263.998 & 161.470 & 305.140 \\
\Block{2-1}{$\dot{v}_{0} \, (m/s^2) $} & mean & 1e-03 & 0 & 0 & 5.8e-03 & \textbf{0} & 0 & 0 & 0 & 0 & \textbf{0}  \\ 
 & max & 1e-03 & 4.145 & 0 & 0.985 & \textbf{0} & 0 & 0 & 0 & 6.9e-01 & \textbf{3.5e-01} \\
\Block{2-1}{$\ddot{v}_{0} \, (m/s^3) $} & mean & 0.391 & 0.554 & 0.353 & 0.043 & \textbf{4.8e-05} & 0.095 & 1.020 & 9.8e-04 & 0.034 & \textbf{0}  \\ 
 & max & 18.217 & 37.817 & 13.778 & 12.167 & \textbf{8.1e-01} & 6.296 & 21.762 & 0.920 & 6.063 & \textbf{1.42} \\
\Block{2-1}{$\dot{\phi}_{0} \, (rad/s) $} & mean & 0 & 2e-04 & 0 & 0 & \textbf{0} & 3.71e-04 & 4.79e-03 & 0 & 6.35e-04 & \textbf{0}  \\ 
 & max & 0 & 0.295 & 0 & 0 & \textbf{0} & 0.385 & 0.795 & 0 & 0.383 & \textbf{0} \\
\Block{2-1}{$\ddot{\phi}_{0} \, (rad/s^2) $} & mean & 2.3e-02 & 5.4e-02 & 2e-02 & 7.33e-03 & \textbf{0} & 0.155 & 0.431 & 6.8e-0.3 & 0.092 & \textbf{2e-03}  \\ 
 & max & 1.290 & 3.242 & 1.210 & 1.003 & \textbf{7e-02} & 2.965 & 6.849 & 1.288 & 3.065 & \textbf{8e-02} \\
\Block{2-1}{$\dot{\theta}_{0} \, (rad/s) $} & mean & 0 & 4e-03 & 0 & 6.86e-03 & \textbf{0} & 0.020 & 0.059 & 2e-03 & 8.06e-03 & \textbf{0}  \\ 
 & max & 1.01e-1 & 4.41e-1 & 5.9e-2 & 0.778 & \textbf{0} & 0.982 & 1.592 & 0.307 & 1.333 & \textbf{0} \\
\Block{2-1}{$\ddot{\theta}_{0} \, (rad/s^2) $} & mean & 0.228 & 0.555 & 0.164 & 0.154 & \textbf{3e-03} & 0.718 & 1.267 & 0.054 & 0.338 & \textbf{1e-03}  \\ 
 & max & 1.941 & 4.863 & 1.704 & 5.003 & \textbf{7e-02} & 7.261 & 14.643 & 2.458 & 7.715 & \textbf{7e-02} \\

\end{NiceTabular}
\normalsize
\caption{\footnotesize{Comparison of $\pi$-MPPI with baseline MPPI schemes for the obstacle avoidance and terrain following scenarios. The table summarizes residuals of constraints on control and its derivatives, along with the success rate and avg. distance to the goal. Lower residuals and avg. distance to goal is better, while a higher success rate is sought.} }
\label{table:obs_and_terrain_base_vs_proj}
\vspace{-0.6cm}
\end{table*}


\subsection{Computation Time}
\vspace{-0.1cm}
\subsubsection{Comparison of $\pi$-MPPI with baseline MPPIwSGF}
Fig.~\ref{fig:comp_time} compares the computation time of $\pi$-MPPI, $\pi$-MPPI-learned, and MPPIwSGF on an RTX3070 i7-11800H laptop. While $\pi$-MPPI takes more time, it still provides $\sim$50 Hz feedback. $\pi$-MPPI-learned, leveraging a neural network for warm-starting, is only slightly costlier than MPPIwSGF in both median and worst-case computation time.
\subsubsection{Comparison of polynomial parametrization with way-point scheme}
One key advantage of polynomial parametrization (Sec. \ref{sec:con_parametrization}) is its ability to reduce the QP dimensionality \eqref{eqn:proj_nu_cost_qp}-\eqref{eqn:proj_ineq_cons_qp}. In contrast to the conventional “way-point” parametrization, where $\overline{\boldsymbol{\nu}}^m$ scales with the planning horizon $K$ (i.e., $3K$ variables), polynomial parametrization represents $\overline{\boldsymbol{\nu}}^m$ and its derivatives with far fewer variables ($3n \ll 3K$). Fig. \ref{fig:poly_time_comparison} compares QP computation times for both methods, showing that as batch size and projection iterations increase, polynomial parametrization remains significantly more efficient.  
\vspace{-0.2cm}

\begin{table}
    \centering
    \small
    \renewcommand{\arraystretch}{1.1} 
    \resizebox{\columnwidth}{!}{%
    \begin{tabular}{|c|c|c|c|c|c|}
        \hline
        \multirow{2}{*}{\textbf{Metrics}} & \multirow{2}{*}{\diagbox[dir=NE]{\textbf{Algorithm}}{\textbf{Num. Proj. Iter.}}} 
        & \multicolumn{2}{c|}{\textbf{Obstacle Avoidance}} 
        & \multicolumn{2}{c|}{\textbf{Terrain Following}} \\
        \cline{3-6}
        & & \textbf{2} & \textbf{5} & \textbf{2} & \textbf{5} \\
        \hline
        \multirow{2}{*}{\% Success} 
        & $\pi$-MPPI-learned & 97.6 & 96.8 & 90.8 & 90.0 \\
        \cline{2-6}
        & $\pi$-MPPI & 94.4 & 99.6 & 82.0 & 98.8 \\
        \hline
        \multirow{2}{*}{Avg. dist. (m)} 
        & $\pi$-MPPI-learned & 131.18 & 128.41 & 264.95 & 257.90 \\
        \cline{2-6}
        & $\pi$-MPPI & 141.93 & 107.99 & 342.43 & 267.25 \\
        \hline
        \multirow{2}{*}{$v_0$ (m/s)} 
        & $\pi$-MPPI-learned & 0.0591 & 0.0559 & 0.5435 & 0.5187 \\
        \cline{2-6}
        & $\pi$-MPPI & 0.0 & 0.9126 & 1.8032 & 1.7457 \\
        \hline
        \multirow{2}{*}{$\phi_0$ (rad)} 
        & $\pi$-MPPI-learned & 0.0035 & 0.0039 & 0.0618 & 0.0333 \\
        \cline{2-6}
        & $\pi$-MPPI & 0.2005 & 0.0810 & 0.1423 & 0.1430 \\
        \hline
        \multirow{2}{*}{$\theta_0$ (rad)} 
        & $\pi$-MPPI-learned & 0.0078 & 0.0081 & 0.0335 & 0.0429 \\
        \cline{2-6}
        & $\pi$-MPPI & 0.3541 & 0.1565 & 0.0325 & 0.0580 \\
        \hline
        \multirow{2}{*}{$\dot{v}_0$ (m/s$^2$)} 
        & $\pi$-MPPI-learned & 0.3717 & 0.3670 & 0.6090 & 0.9097 \\
        \cline{2-6}
        & $\pi$-MPPI & 0.0 & 0.3129 & 0.1890 & 1.7015 \\
        \hline
        \multirow{2}{*}{$\dot{\phi}_0$ (rad/s)} 
        & $\pi$-MPPI-learned & 0.0 & 0.0 & 0.0 & 0.0 \\
        \cline{2-6}
        & $\pi$-MPPI & 0.0950 & 0.0922 & 0.2329 & 0.1505 \\
        \hline
        \multirow{2}{*}{$\dot{\theta}_0$ (rad/s)} 
        & $\pi$-MPPI-learned & 0.0 & 0.0 & 0.0 & 0.0 \\
        \cline{2-6}
        & $\pi$-MPPI & 0.2264 & 0.0757 & 0.0 & 0.0167 \\
        \hline
        \multirow{2}{*}{$\ddot{v}_0$ (m/s$^3$)} 
        & $\pi$-MPPI-learned & 1.8406 & 1.7366 & 3.1984 & 2.7405 \\
        \cline{2-6}
        & $\pi$-MPPI & 0.0 & 1.0179 & 1.9961 & 3.9043 \\
        \hline
        \multirow{2}{*}{$\ddot{\phi}_0$ (rad/s$^2$)} 
        & $\pi$-MPPI-learned & 0.0944 & 0.0889 & 0.3811 & 0.2762 \\
        \cline{2-6}
        & $\pi$-MPPI & 0.4217 & 0.4160 & 0.6124 & 0.5501 \\
        \hline
        \multirow{2}{*}{$\ddot{\theta}_0$ (rad/s$^2$)} 
        & $\pi$-MPPI-learned & 0.2088 & 0.2107 & 0.2890 & 0.2591 \\
        \cline{2-6}
        & $\pi$-MPPI & 0.5539 & 0.4776 & 0.1670 & 0.2267 \\
        \hline
    \end{tabular}
    } 
    \caption{\footnotesize{Comparison of $\pi$-MPPI and $\pi$-MPPI-learned for Obstacle Avoidance and Terrain Following scenarios, showing success rate, average distance to the goal, and maximum constraint residuals.}}
    \label{tab:leanred_vs_proj_comparison}
    \vspace{-0.4cm}
\end{table}

\begin{table}
    \centering
    \renewcommand{\arraystretch}{1.0}
    \small
    \begin{adjustbox}{max width=0.98\linewidth}
    \begin{tabular}{|l|c|c|c|c|c|c|}
        \hline
        \multirow{2}{*}{Metrics} & \multicolumn{3}{c|}{Obstacle Avoidance} & \multicolumn{3}{c|}{Terrain Following} \\
        \cline{2-7}
        & \shortstack{Neural warm-\\start} & \shortstack{$\boldsymbol{\nu}^m$-\\init.} & \shortstack{Direct\\Sol.} & \shortstack{Neural warm-\\start} & \shortstack{$\boldsymbol{\nu}^m$-\\init.} & \shortstack{Direct\\Sol.} \\
        \hline
        $v_0$ (m/s) & \textbf{0.0591} & 1.5625 & 5.7907 & \textbf{0.5435} & 1.9395 & 21.3833 \\
        \hline
        $\phi_0$ (rad) & \textbf{0.0035} & 0.2119 & 0.1908 & \textbf{0.0618} & 0.2563 & 0.2944 \\
        \hline
        $\theta_0$ (rad) & \textbf{0.0078} & 0.2801 & 0.2644 & \textbf{0.0335} & 0.1362 & 0.2586 \\
        \hline
        $\dot{v}_0$ (m/s$^2$) & \textbf{0.3717} & 0.4098 & 3.9304 & \textbf{0.6090} & 1.8993 & 8.2567 \\
        \hline
        $\dot{\phi}_0$ (rad/s) & \textbf{0.0} & 0.1661 & 0.2196 & \textbf{0.0} & 0.1969 & 0.3522 \\
        \hline
        $\dot{\theta}_0$ (rad/s) & \textbf{0.0} & 0.2033 & 0.1976 & \textbf{0.0} & 0.0146 & 0.2332 \\
        \hline
        $\ddot{v}_0$ (m/s$^3$) & 1.8406 & 1.2385 & 3.0118 & \textbf{3.1984} & 5.6318 & 8.4345 \\
        \hline
        $\ddot{\phi}_0$ (rad/s$^2$) & \textbf{0.0944} & 0.6066 & 0.5016 & \textbf{0.3811} & 0.7405 & 0.7159 \\
        \hline
        $\ddot{\theta}_0$ (rad/s$^2$) & \textbf{0.2088} & 0.6014 & 0.6137 & \textbf{0.2890} & 0.3166 & 0.9369 \\
        \hline
    \end{tabular}
    \end{adjustbox}
    \caption{\footnotesize{Comparison of accelerated $\pi$-MPPI schemes (Neural warm-start, $\boldsymbol{\nu}^m$-initialization, Direct solution) for Obstacle Avoidance and Terrain Following scenarios, showing maximum constraint residuals.}}
    \label{tab:learned_initialization_comparison}
    \vspace{-0.8cm}
\end{table}

\section{Conclusions and Future Work} \label{sec:conclusion}
We introduced $\pi$-MPPI, which integrates a QP-based projection filter within the MPPI pipeline to enforce arbitrary bounds on controls and their derivatives. This makes it ideal for agile systems like FWVs, where smooth controls prevent oscillations and actuator stalls. Additionally, $\pi$-MPPI requires minimal tuning of the control noise covariance and tolerates large perturbations, enabling better exploration and task completion than MPPIwSGF. Finally, we demonstrated that a neural network trained in a self-supervised manner accelerates the convergence of our custom QP solver. 

In the future, we seek to apply $\pi$-MPPI for applications like autonomous driving, where smooth controls can directly impact ride comfort. We are also looking to embed a non-convex optimizer within MPPI that can project not only control but also state constraints onto a feasible set. 
\vspace{-0.3cm}

\bibliographystyle{IEEEtran}
\bibliography{ref}


\end{document}